\documentclass[times,twocolumn,final]{elsarticle}

\graphicspath{{./figures/}}
\usepackage{xr}
\usepackage{graphicx}
\usepackage{natbib}
\usepackage{xcolor}
\usepackage{booktabs}
\usepackage{multirow}
\usepackage{textgreek}
\usepackage{subfig} 
\usepackage{enumitem}
\usepackage{url}
\usepackage{hyperref}
\usepackage[utf8]{inputenc}
\usepackage{array}
\usepackage{url}
\biboptions{authoryear}
\usepackage[none]{hyphenat}

\begin{document}
\begin{frontmatter}
\title{Navigating the landscape of multimodal AI in medicine: a scoping review on technical challenges and clinical applications}%

\author[1,7]{Daan Schouten\fnref{joint}\corref{cor1}}
\author[2,3]{Giulia Nicoletti\fnref{joint}}
\author[4,1]{Bas Dille\fnref{joint}}
\author[1,5,6]{Catherine Chia}
\author[1]{Pierpaolo Vendittelli}
\author[1]{Megan Schuurmans}
\author[1,7]{Geert Litjens}
\author[1]{Nadieh Khalili}

\cortext[cor1]{Corresponding author. (\textit{email:} Daan.Schouten@radboudumc.nl)}
\fntext[joint]{These authors contributed equally to this work}
\address[1]{Department of Pathology, Research Institute for Medical Innovation, Radboud University Medical Center, Nijmegen, the Netherlands}
\address[2]{Department of Electronics and Telecommunications, Polytechnic University of Turin, Turin, Italy}
\address[3]{Department of Surgical-Medical Sciences, Unversity of Turin, Turin, Italy}
\address[4]{Department of Radiology and Nuclear Medicine, Erasmus University Medical Centre, Rotterdam, the Netherlands}
\address[5]{Department of Dermatology, Erasmus University Medical Centre, Rotterdam, the Netherlands}
\address[6]{Department of Pathology, Erasmus University Medical Centre, Rotterdam, the Netherlands}
\address[7]{Oncode Institute, Utrecht, the Netherlands}


\begin{abstract}
Recent technological advances in healthcare have led to unprecedented growth in patient data quantity and diversity. While artificial intelligence (AI) models have shown promising results in analyzing individual data modalities, there is increasing recognition that models integrating multiple complementary data sources, so-called multimodal AI, could enhance clinical decision-making. This scoping review examines the landscape of deep learning-based multimodal AI applications across the medical domain, analyzing 432 papers published between 2018 and 2024. We provide an extensive overview of multimodal AI development across different medical disciplines, examining various architectural approaches, fusion strategies, and common application areas. Our analysis reveals that multimodal AI models consistently outperform their unimodal counterparts, with an average improvement of 6.2 percentage points in AUC. However, several challenges persist, including cross-departmental coordination, heterogeneous data characteristics, and incomplete datasets. We critically assess the technical and practical challenges in developing multimodal AI systems and discuss potential strategies for their clinical implementation, including a brief overview of commercially available multimodal AI models for clinical decision-making. Additionally, we identify key factors driving multimodal AI development and propose recommendations to accelerate the field's maturation. This review provides researchers and clinicians with a thorough understanding of the current state, challenges, and future directions of multimodal AI in medicine. 
\end{abstract}


\end{frontmatter}

\section{Introduction}
The healthcare landscape is evolving rapidly, driven by an increasingly data-centric approach to patient care and decision-making \citep{shilo_axes_2020}. This shift is complemented by the advent of technologies such as digital pathology \citep{niazi_digital_2019}, biosensors \citep{sempionatto_wearable_2022}, and next-generation sequencing \citep{steyaert_multimodal_review_2023}, which provide clinicians with novel insights in various domains. The data generated by these diverse modalities is generally complementary, with each modality contributing unique information to the status of a patient. Some modalities offer a comprehensive overview at the macro level, while others may provide detailed information at single-cell resolution \citep{steyaert_multimodal_review_2023}. In addition to this recent growth in data quantity, there is a concurrent increase in the quality and diversity of available treatment options. Hence, selecting the optimal treatment has become increasingly complex, and a further data-centric approach to treatment selection may be required.

The traditional approach to integrating information from different data modalities into a single decision is represented by multidisciplinary boards, where each specialized clinician offers their perspective on a given modality or piece of information in pursuit of consensus \citep{mano_implementing_2022}. Although establishing these boards has improved disease assessments and patient management plans \citep{mano_implementing_2022}, there is a foreseeable limit to the scalability of these boards. If data quantity and diversity continue to rise, many domain experts will be required to integrate these different information streams effectively. Fortunately, another technological advancement that is gaining a foothold in healthcare is artificial intelligence (AI). Although the vast majority of published work focuses on single modality applications of AI, several authors have highlighted the potential of AI systems to combine multiple streams of information, so-called multimodal AI, for decision-making \citep{steyaert_multimodal_review_2023, acosta_multimodal_2022, lipkova_artificial_2022}. These multimodal AI models are trained to process different streams of multimodal data effectively, leverage the complementary nature of information, and make an informed prediction based on a broader context of the patient's status. However, despite these promising results, studies investigating multimodal AI models are comparatively scarce, and the development of unimodal models remains the de facto standard.

This lagging development of multimodal AI models can be attributed to several challenges. First, a practical challenge can be found in the cross-departmental nature of multimodal AI development. As different data modalities may originate from various medical departments, consulting different medical domain experts will likely be required for effective data integration. In addition, medical departments may have varying experience in data storage, retrieval, and processing, limiting the possibilities of multimodal AI development. For example, if a radiology department has a fully digital workflow while the corresponding pathology department does not, this effectively prohibits multimodal AI endeavors where whole slide images would be combined with radiological imaging data. 

Different data modalities can have vastly different characteristics, such as dimensionality or color space, which generally requires different AI model architectures tailored towards those modalities, increasing model design complexity. For example, convolutional neural networks (CNN) were initially proposed for structured data, such as 2D and 3D images, but can't straightforwardly be applied to unstructured data. Conversely, transformers are solid, flexible encoders for various data modalities. Still, whether a one-size-fits-all architecture can capture various medical data modalities effectively remains unclear. In practice, multimodal data integration is commonly achieved using different (intermediate) model outputs. Training multiple domain-specific AI models (i.e., encoders) and efficiently integrating these in a single prediction poses a challenge unique to multimodal AI development. 

Last, the inconsistent availability of all modalities for each patient within a multimodal dataset adds complexity. Patients with different disease trajectories will have various available modalities, leading to partially incomplete datasets. This can substantially reduce the adequate training dataset size for AI models that require complete multimodal data to generate predictions. Moreover, these issues also translate to implementation. If modalities are missing, it may be unclear how this impacts the model's performance from the perspective of fewer available data to base a decision on and the potential introduction of population selection bias \citep{acosta_multimodal_2022}. In short, developing multimodal AI models poses several novel challenges compared to unimodal AI development.

Even given these challenges, several works have been done in the past on multimodal AI applications, typically involving handcrafted features. A key issue with these approaches was that the difficulties requiring particular domain expertise are multiplied, as expert clinicians would also need to be involved in the feature design phase \citep{vaidya_ct_2020, tortora_radiopathomics_2023}. An excellent overview was published by \citet{kline_multimodal_2022}, indicating that these models obtained a 6.4\% mean improvement in AUC compared to their unimodal counterparts. 

Recent years have shown an accelerated interest in multimodal AI development for medical tasks \citep{salvi_multi-modality_2024}, as using unsupervised learning and deep neural networks as encoders has significantly simplified the feature extraction step. In this review, we comprehensively summarize the state-of-the-art in multimodal AI development for medical tasks and investigate to what extent multimodal data integration is living up to its purported benefits. Unlike previous reviews that have focused on specific diseases, prediction tasks, or modality combinations \citep{acosta_multimodal_2022, salvi_multi-modality_2024, KRONES2025102690}, our analysis encompasses the full spectrum of the medical domain. Specifically, our review aims to shed light on I) the progress of multimodal AI model development across different medical disciplines and tasks, II) the technical challenges inherent in multimodal AI development, including model architectures, fusion methods, and the handling of missing data, III) the foreseeable road to the clinic of multimodal AI models, addressing aspects such as regulatory approval and explainability, and IV) the factors driving multimodal AI development and potential strategies to promote further maturation of this field. Lastly, we will provide an outlook on future perspectives in multimodal AI development based on our careful analysis of 432 papers published over the past six years (2018-2024).

\section{Search Criteria} \label{ssec:Search}
This scoping review aimed to evaluate the application of multimodal AI models in the medical field, where we define multimodality as data originating from different medical specialties. For instance, we consider a model to be multimodal when it integrates a diagnostic CT scan and subsequent tissue biopsy slides, as it combines the domains of the radiologist and the pathologist, respectively. Conversely, a model integrating T1- and T2-weighted MRI scans would not be considered multimodal. In addition to this multimodality criterion, we limited our scope to I) studies using deep neural networks and II) studies developing multimodal models for specific medical tasks (i.e., no generic visual question answering).

\begin{figure}
    \centering
    \includegraphics[width=\linewidth]{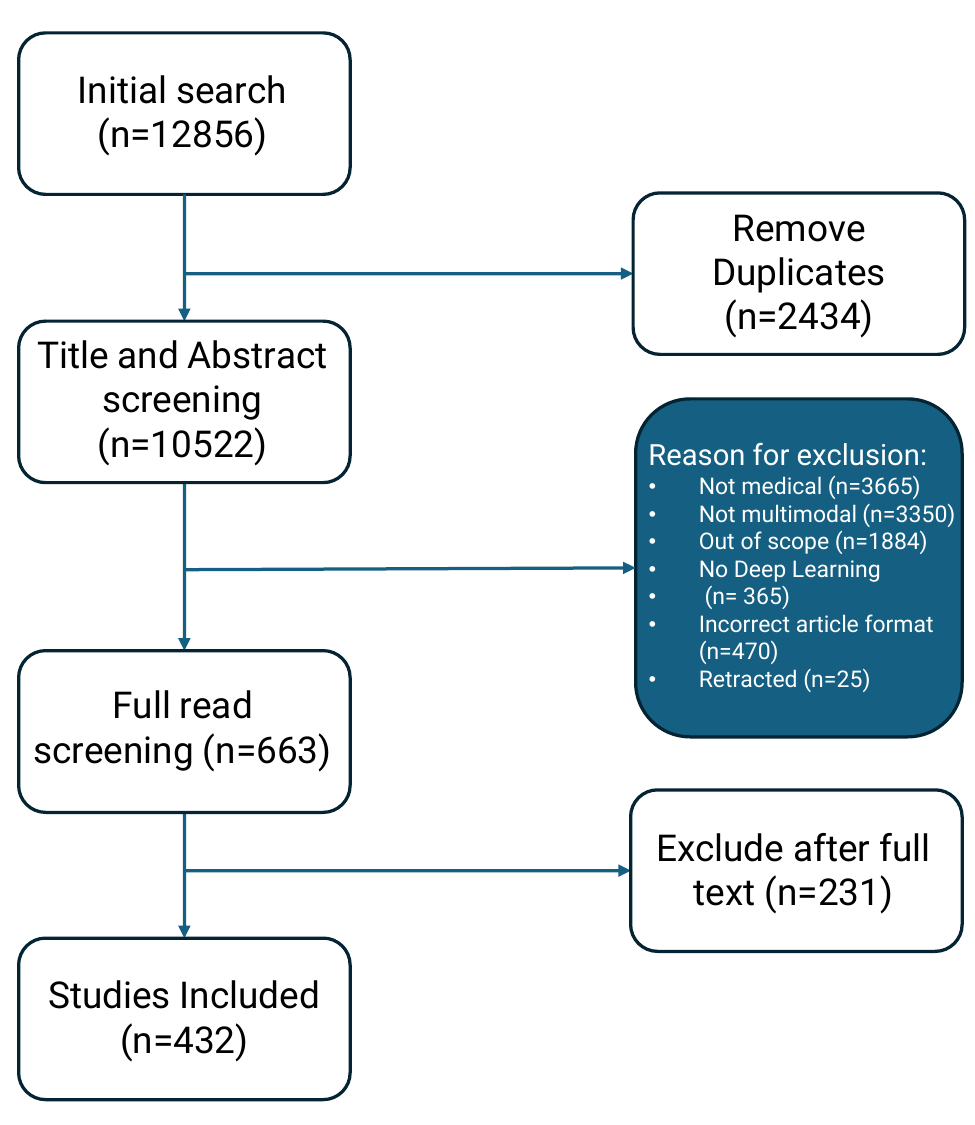}
    \caption{Overview of the screening process.}
    \label{fig:screening}
\end{figure}

The literature search was conducted in PubMed, Web of Science, Cochrane, and Embase. The entire search string per database can be found in the supplementary materials. The search was initially performed on April 16, 2024 and repeated on October 2, 2024 to ensure an up-to-date overview, yielding a total of 12856 initial results. These results included both journal and conference papers. In addition to the previously mentioned inclusion criteria, we applied additional standard exclusion criteria summarized in Figure \ref{fig:screening}. Specifically, we excluded review articles, non-English publications, articles without full text, non-peer-reviewed preprints, and those published before 2018. After filtering with these criteria and deduplicating results using the DedupEndNote tool \citep{Lobbestael2023}, 10522 articles were imported into the paper screening software Rayyan \citep{Ouzzani2016} for the Title and Abstract (TiAb) screening phase.
A team of six reviewers conducted the TiAb screening phase. Each paper's title and abstract were reviewed based on the inclusion criteria. Included papers were verified by a second reviewer, and potential discrepancies were resolved through discussion or the involvement of a third reviewer if necessary. The TiAb screening phase was performed sequentially, starting with verifying that the paper is in the medical domain, then determining whether it is in scope, whether multimodality is involved, and finally, assessing whether deep neural networks are used. Papers were deemed to be outside the scope of this review when they investigated tasks that do not have an explicit clinical question (i.e., image denoising, registration). The TiAb screening resulted in 663 articles being considered for full-text screening.

In the full-text reading phase, a single reviewer read each article in full to confirm adherence to our inclusion criteria. In case of doubt, a second reviewer was involved to arrive at a consensus decision. Eventually, 432 studies were included in the final analysis (see the supplementary materials for the entire list of included papers).

\section{Overview of multimodal medical AI}\label{overview_modalities}
The landscape of multimodal artificial intelligence (AI) in medical research has expanded between 2018 and 2024, as shown by the growing number of articles focused on integrating multiple modalities. Subdividing the 432 articles in this review per year (Figure \ref{fig:fig_distribution_of_modalities}a), we see a rapid increase, starting with 3 papers in 2018 to 150 papers in 2024 at the end of the data collection of this review.

We distributed the reviewed papers according to data modalities, which will be broadly described here. Subsequently, the review is organized according to the following subsections:

\begin{itemize}
    \item State-of-the-art algorithm designs for multimodal medical AI (section \ref{sssec:Methodological approaches})
    \item Clinical tasks (section \ref{sssec:Clinical applications}) for a detailed analysis of the multimodal AI approaches categorized per organ system
    \item Challenges and opportunities for clinical adoption (section \ref{ssec:road_to_the_clinic}) of multimodal AI
\end{itemize}

Last, we will conclude with a high-level discussion of our findings and provide guidelines for future research directions.

\subsection{Modalities and data types}
We divided the data modalities into image-based and non-image-based modalities. The image-based modalities are grouped according to their related medical specialties. This resulted in the following categories: radiology (computed tomography (CT), magnetic resonance imaging (MRI), ultrasound (US), X-rays, and nuclear imaging (SPECT/PET)), pathology (stained histology images) and 'clinical images' (including optical coherence tomography (OCT), fundus photography, and dermatoscopy, among others). If there were too few papers for a medical specialty, we grouped them into a catch-all 'other images', a subset of 'clinical image' category. Meanwhile, the non-image-based modalities consist of text (structured text such as tabular laboratory results and unstructured text such as free-text medical reports), omics data (e.g., genomics, transcriptomics, or proteomics), and other non-image modalities (such as Electroencephalography (EEG) or Electrocardiography (ECG) signals).

\begin{figure*}%
    \centering
    \includegraphics[width=19cm]{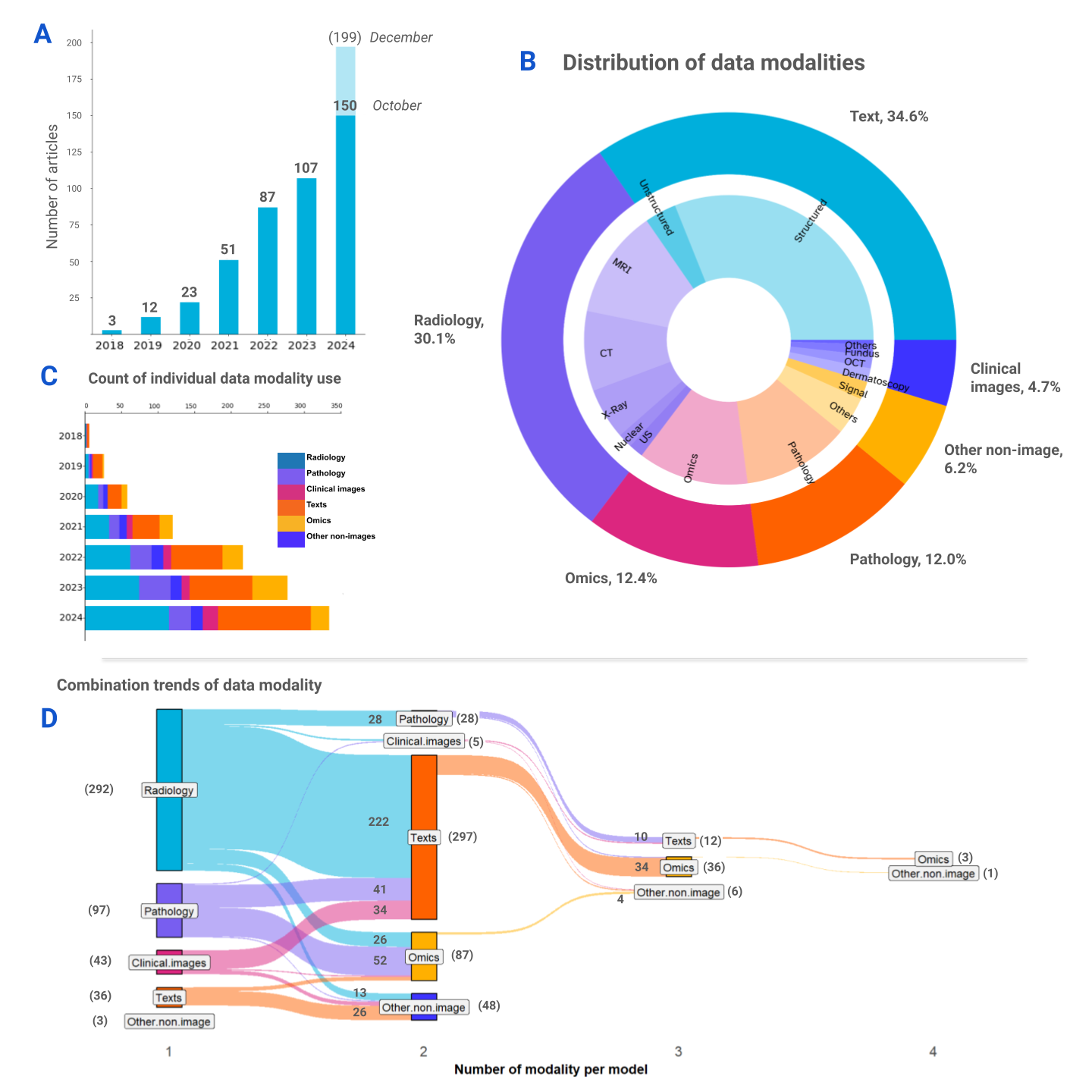}%
    \caption{Overview of the data modalities used in the reviewed articles. (A) Distribution of articles by year. Bar chart shows an exponential increment in the number of studies per year from 2018 to 2024. Extrapolating, the number of multimodal medical AI studies is expected to reach 199 by the end of 2024. (B) Pie chart shows the proportions of different modality groups and the respective data modalities used across studies. (C) Stacked bar chart illustrates the growth trends of data modality groups over the years. Note that the values used in this chart represent the counts of individual data modality uses, where multiple modalities could be presented in a single article. (D) Diagram shows the combination trends between data modalities per model. The diagram captures the unique modality combinations presented in each models the individual article has presented. The numbers in brackets indicate the total summation of models per category, whereas the numbers without brackets represent the count of models of each combination, visualized with the ribbon bands between the vertical nodes. The majority of the models used two data modalities, and a portion of the total used three and four modalities. Three multimodal models used data modalities that were grouped under "other non-image" category based on the definition used in this review. \\}%
    \label{fig:fig_distribution_of_modalities}%
\end{figure*}

Radiology and text were the most commonly used modalities (each 30\%), followed by omics (12\%) and pathology (12\%). Figure \ref{fig:fig_distribution_of_modalities}b visualizes all modalities and their respective subtypes. Over the years, the trend of using radiology and text modalities remained the most common (see Figure \ref{fig:fig_distribution_of_modalities}c). The most prevalent combination is radiology with text (n=206), followed by pathology/omics (n=51), clinical images/text (n=33), radiology/omics (n=24), pathology/text (n=22), and radiology/pathology (n=16). These combinations showcase a preference for integrating one imaging modality with structured or unstructured text data. A complete overview is presented in Figure \ref{fig:fig_distribution_of_modalities}d.

The review also reveals several complex combinations involving three or more modalities, albeit in smaller numbers. Notably, pathology/text/omics (n=19), radiology/text/omics (n=15), radiology/pathology/text (n=7), and radiology/pathology/omics/text (n=3) demonstrate the attempts to create comprehensive models that span multiple scales of biological organization - from organ-level (radiology) to tissue and cellular (pathology) and subcellular (omics), complemented by clinical context (text). An exhaustive overview of all modality combinations can be found in the supplementary materials.

\subsection{Organ systems, medical tasks and AI functions} \label{sssec:Key application areas}
\begin{figure*}%
    \centering
    \includegraphics[width=20cm]{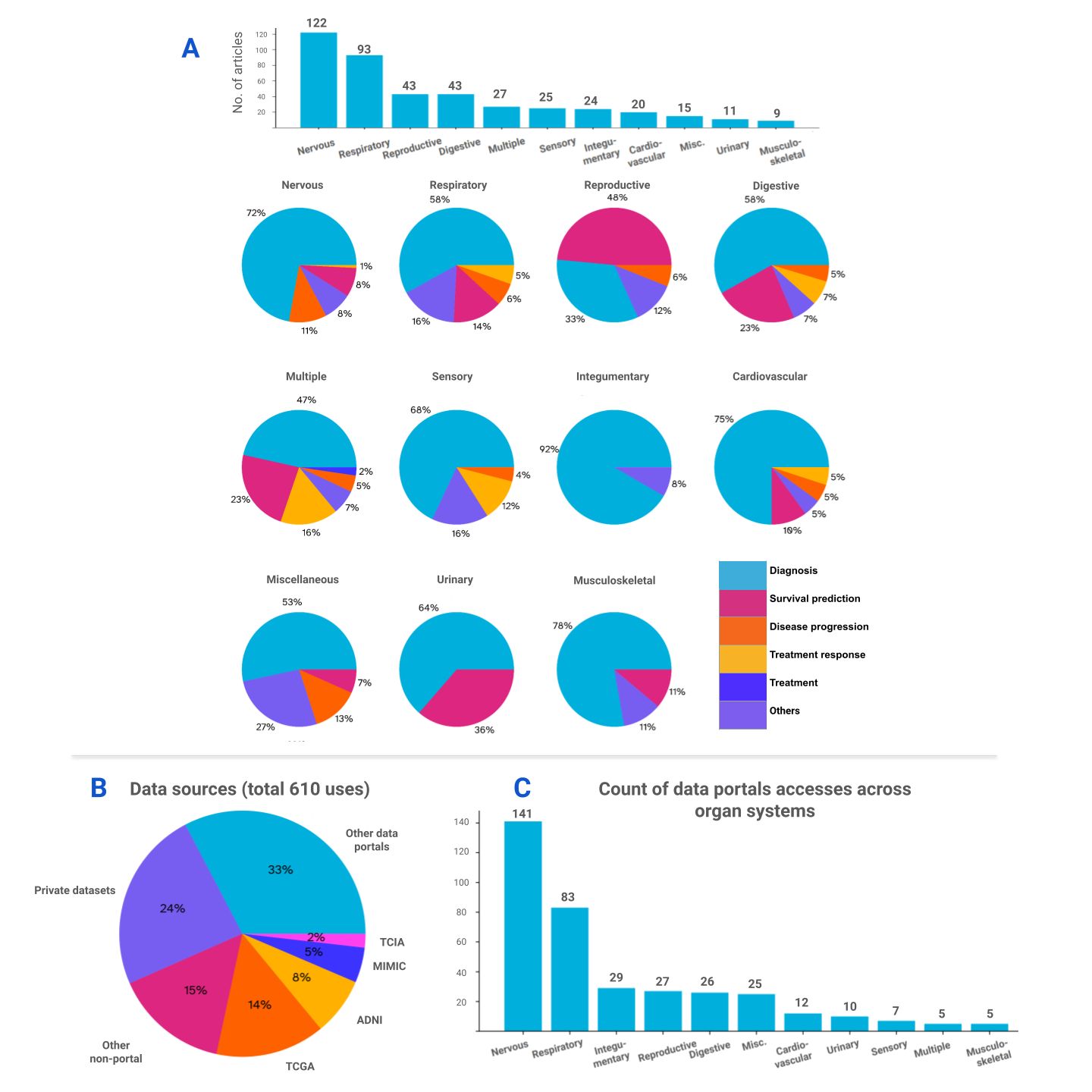}%
    \caption{A deeper dive into the medical tasks and data sources of the review. The numbers on the bars indicate the total summation per category. (A) Top: The number of articles per organ system. Bottom: Distribution of medical tasks across organ systems. Pie charts show diagnosis being the most prevalent medical task performed in studies of all organ systems. (B) The use trends of data sources in this review. Note that the values used in the chart represent the total count of uses of all the reviewed studies, where multiple data sources could be referred to in each study. About 61\% of the total uses were sourced from data portals (e.g. TCGA, ADNI, etc.), 15\% from research data shared publicly by publications, and 24\% of the data uses were private datasets that were not made public. (C) Distribution of public data sources (excluding private datasets) across the studies of organ systems. Similarly, the nervous and respiratory systems are leading in the count of public data uses. A detailed breakdown of these public data sources can be found in the supplementary materials.\\}%
    \label{fig:fig_distribution_medical_disciplines_tasks}%
\end{figure*}

We categorized all studies into eleven organ systems (see Figure \ref{fig:fig_distribution_medical_disciplines_tasks}a for the complete list). The nervous system dominates with 122 studies, followed by respiratory (n=93), reproductive (n=43), digestive (n=43), sensory (n=25) and integumentary (n=24). A substantial number of studies (n=15) fall under the miscellaneous category and 27 studies involved multiple organ systems, which are further explained in the dedicated section Clinical applications (section \ref{sssec:Clinical applications}).

Six medical task categories: diagnosis, estimating prognosis (which is further subdivided into survival prediction, disease progression, and treatment response analyses), treatment, and others. Diagnosis emerged as the primary focus, accounting for 45\% (multiple systems) to 91\% (integumentary system) of the medical tasks across all organ systems (see Figure \ref{fig:fig_distribution_medical_disciplines_tasks}a and b). Survival prediction was the second most common task (18\% of all medical tasks). Other tasks, such as prediction of disease progression or treatment response, were not uniformly represented across all organ systems.

\section{Methodology} \label{sssec:Methodological approaches}
\subsection{Importance of public data}
As stated in the introduction, data availability is a key challenge for the development of multimodal medical AI. This is why we see a strong correlation between the number of models for a specific organ system/modality combination and the availability of public data (see Figure \ref{fig:fig_distribution_medical_disciplines_tasks}c). The utilization of publicly shared datasets in multimodal AI research for medical applications is widespread, with 61\% of the data sources used in the model development coming from public data portals such as The Cancer Genome Atlas (TCGA, 14\%), Alzheimer's Disease Neuroimaging Initiative (ADNI, 8\%), Medical Information Mart for Intensive Care (MIMIC, 5\%) and The Cancer Imaging Archive (TCIA, 2\%), 15\% from data shared publicly through other means (e.g. GitHub, publisher's website), and 24\% from private datasets which were not shared publicly. We grouped all other data portals used by less than ten reviewed papers into "other data portals" (20\%). A detailed breakdown of these public data sources can be viewed in the supplementary materials.

\subsection{Feature encoding and modality fusion}
The advent of deep neural networks was a key accelerator for multimodal medical AI. These networks significantly simplified the feature extraction/encoding step for individual modalities. Each modality could now be encoded by its deep neural network, and the resultant features could be combined for downstream tasks. Powerful self-supervised learning techniques such as DINO \citep{caron2021emerging} and SimCLR \citep{chen2020simple} have now also enabled the training of these feature encoders without any labels, further increasing the attractiveness of this approach. However, we have seen that there is still significant diversity in approaches across the reviewed papers, which we will detail in the following subsections.

Various encoder mechanisms are used for different data modalities. Encoders are categorized into Convolutional Neural Networks (CNNs), Vision Transformers (ViTs), Recurrent Neural Networks (RNNs), Transformers, handcrafted feature encoders, multi-layer perceptrons (MLPs), multiple encoders, and 'other' (e.g. graph neural networks). 
Our analysis reveals that CNNs dominate the encoder landscape (used in 82\% of the studies), followed by 'other' (32\%), and MLPs (21\%). Unsurprisingly, CNNs show strong correlations with image-based data modalities, including radiology, pathology, and other (miscellaneous) image modalities. In contrast, non-image modalities, such as 'omics and (un)structured text, use a more diverse range of encoders, including handcrafted feature encoders, MLPs, RNNs, and Transformers. 

\begin{figure*}%
    \centering
    \includegraphics[width=15cm]{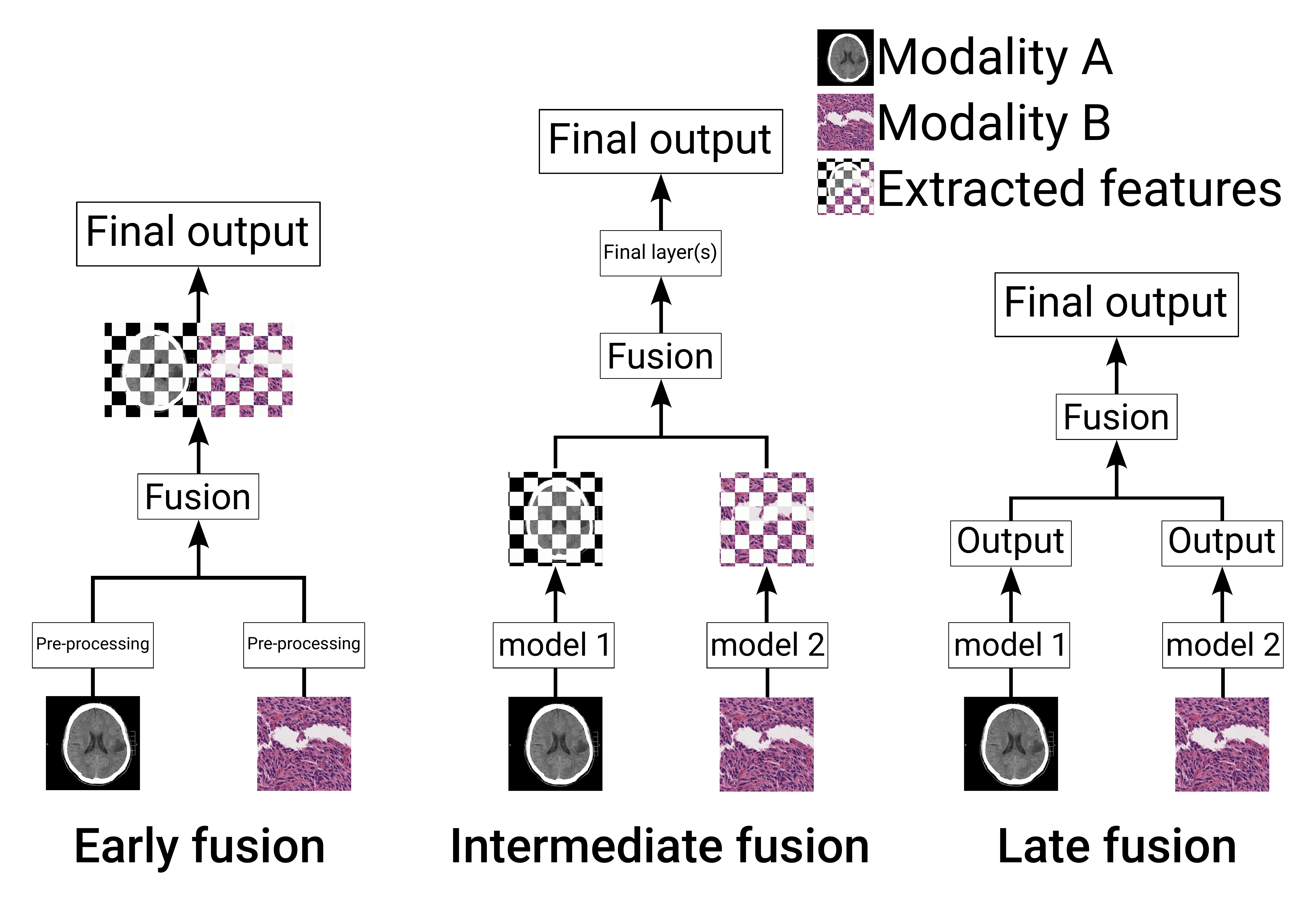}%
    \caption{Simplified 
    schematic view of the different fusion stages.}%
    \label{fig:Multimodal_Fusion}%
\end{figure*}

The second important design decision is how and when the differing modalities are fused. The fusion stage can vary depending on the model structure, which often dictates the optimal point for integrating information across modalities. A schematic view of these fusion stages is shown in Figure \ref{fig:Multimodal_Fusion}.

Out of all papers reviewed in this study, the vast majority (79\%, 341/432 papers) utilized intermediate fusion, in which data sources get fused after feature encoding but before the final layers of the neural network (e.g., the classification or regression head). A common strategy is simply concatenating the feature vectors of the different unimodal modality encoders and feeding the resultant vector to the final layers. This concatenation method was used in most models (69\%) using intermediate fusion, and was not limited to certain modality combinations or clinical application areas \citep{lee_predicting_2019, zhi_multimodal_2022, liu_multi-task_2024}. However, several studies found through ablation experiments that applying other methods to fuse the feature vectors (i.e. taking the outer product, Kronecker product or compact bilinear pooling), outperformed concatenation by a notable margin \citep{yang_prediction_2022, wang_predicting_2023, wang_multi-modality_2024_fuse}. Another common intermediate fusion technique (12\%) was the use of attention, where the unimodal embeddings were passed through an attention mechanism to optimally learn from the complementary information in both embeddings \citep{kayikci_breast_2023, liu_cascaded_2023, machado_reyes_identifying_2024}.

Late fusion, a method in which fusion is performed by combining results or predictions of unimodal models, is the second most common technique (14\%). These architectures can look similar to those from intermediate fusion models, but without intermediate components such as attention mechanisms or mixing layers. However, most of these late fusion approaches combine unimodal models and their individual predictions to get a multimodal result. This was often (32\%) achieved by applying a (weighted) average over the predictions for each modality \citep{ying_multi-modal_2021, caruso_multimodal_2022, jung_multimodal_2024}) or by training a separate model on top of the unimodal predictions (37\%). The latter was mainly done through regression or traditional machine learning models like random forests, boosting algorithms or Cox models \citep{ma_brain_2020, kolk_multimodal_2024, wang_deep_2024}. A key aspect of late fusion is that no interaction exists between the modalities in the learning process, meaning that the training data for the respective unimodal models does not have to be paired. It therefore becomes easier to handle missing data in these models, as patients with only one modality can be used to train only the model of that modality without the need to infer the missing ones. On the contrary, the lack of interaction could potentially limit model expressiveness.

The last method involves early fusion, at which modalities are fused before feature encoding. Our study found that this fusion method has been applied the least (6\%). A key challenge is that the input data has to exist in the same 'space' to allow early fusion. The difficulty beyond this kind of fusion could be influenced by the types of modalities involved in the fusion, for example, combining radiology and pathology images might often require some form of image registration, which is an unsolved challenge for many clinical applications. However, several early fusion methods were implemented that do not require extensive preprocessing steps, ranging from a simple concatenation of the modalities \citep{shi_attention-based_2023, lopez_reducing_2020} to more complex representations involving graph networks \citep{lei_alzheimers_2024}, recurrent neural networks \citep{xu_multi-modal_2022}, and cross-modality representation alignment networks \citep{wu_camr_2023}. Other interesting methods combined image and non-image modalities by directly marking \citep{pelka_sociodemographic_2020} or multiplying \citep{qiu_towards_2024} clinical variables onto images, thereby enhancing the image data with relevant contextual information. A key potential advantage of early fusion is that modalities are already available in the feature encoding stage, potentially allowing deep neural networks to optimize feature design by leveraging all information simultaneously.

\subsection{Network architectures} \label{sssec:architecture choices}
The architecture choices for multimodal AI models often rely on the purpose(s) of the model and data availability. Regardless, the critical steps when designing such a model include the same core functionalities in feature extraction, information fusion, and final processing of the fused information. Some papers also introduce explainability \citep{ketabi_multimodal_2023, yang_deep_2021, parvin_convolutional_2024} into their models or attempt to construct multiple different model architectures to assess the effectiveness of each variant \citep{huang_multimodal_2020, lopez_reducing_2020, caruso_multimodal_2022}. 
Examining the usage of individual modality encoders over the years reveals that contemporary feature extraction models like (vision) transformers are gaining popularity, while on the other hand, MLPs have seen a decrease in usage, with CNN usage staying roughly the same in terms of popularity over the last few years. Traditional machine learning methods are also still employed for feature encoding in 5\% of all papers, often to extract features from structured text data.

\subsection{Handling missing modalities} \label{sssec:Missing data handling}
Most multimodal AI models work on the assumption of data completeness: all modalities must be available for every entry. However, this is often not feasible due to issues such as data silos caused by incompatible data archival systems (i.e. PACS), a study's retrospective nature, or simply data privacy concerns. Handling missing data without introducing bias into the analysis presents a significant challenge. Indeed, a commonly employed approach to address this issue is to exclude all entries with at least one missing modality and include only complete entries in the dataset used for analysis, and this is the case for 69\% of the included papers. Although the exclusion of incomplete entries is a quick and straightforward method for handling missing data, this approach results in a reduction in data sample size with a subsequent loss of potentially relevant samples. Furthermore, this limits the model inference to complete-modality data and inevitably leads to selection biases, thereby potentially compromising validity and generalizability. Rather than discarding incomplete samples, missing information can also be estimated using imputation techniques, thereby alleviating both aforementioned disadvantages of fully excluding incomplete entries. These data imputation techniques can be broadly divided into non-learning-based and learning-based approaches.

\subsubsection{Non-learning-based approaches}
The findings of our study indicate that non-learning-based approaches are commonly utilized, occurring in 45\% (35/78) of all papers that reported any method to account for missing data. These techniques are primarily employed to impute values of structured data (n=29), such as clinical variables and test results. Less frequently, these methods are used to address missing gene values (n=2) or image modalities (n=2). For continuous variables, imputation is performed using measures of central tendency such as the mean or median, or performing a moving average, which replaces missing values with the average of a specified number of surrounding data points. For categorical variables, imputation often involves using the mode or adding a new category for missing values. Some studies apply fixed values, like zeros, -1, or random values from similar records (hot-deck method). Although these methods are simple and easy to implement, they can introduce bias and reduce data variability \citep{flores_missing_2023}.

\subsubsection{Learning-based approaches}
Some studies leveraged traditional machine learning methods to predict missing values, learning patterns from complete data as an alternative to simpler imputation techniques. Common methods included k-nearest neighbor (k-NN), which imputes missing values based on the values of the nearest neighbors \citep{ross_external_2024, qiu_multimodal_2022, lee_multimodal_2024_gen}, and the weighted nearest neighbor approach \citep{wu_camr_2023, kayikci_breast_2023, mustafa_ensembled_2023, palmal_integrative_2024}, which assigns weights to neighbors based on their distance. Other techniques used include linear regressions \citep{ghafoori_predicting_2023}, random forests \citep{yu_transformer-based_2024, kolk_multimodal_2024}, neural networks \citep{menegotto_computer-aided_2021} and Classification and Regression Trees (CART) algorithm \citep{yin_development_2024}, as well as advanced methods like Multivariate Imputation by Chained Equations (MICE) algorithm \citep{rahman_bio-cxrnet_2023, liu_preoperative_2022, lim_use_2022, kim_multimodal_2024} and XGBoost \citep{feher_artificial_2024, zambrano_chaves_opportunistic_2023, fan_multimodal_2024}, which handle missing data by respectively incorporating multiple imputations and tree-based learning during training. \\
Other methods applied deep learning models to impute missing values or manage an arbitrary number of modalities, directly modifying the model architecture to process incomplete data without imputation. Some deep-learning-based imputation strategies leverage different generative models, such as auto-encoders \citep{xu_explainable_2022, akramifard_early_2021}, or generative adversarial networks \citep{dolci_deep_2023}. Others directly predict the missing data at the output layer \citep{saad_learning-based_2022} or at previous visits of a Recurrent Neural Network \citep{xu_multi-modal_2022}. However, there seems to be little consensus or evidence to suggest that one method should be preferred. Some papers have addressed the missing data challenge by introducing specific drop-out modules to train on or simulate missing data \citep{cheerla_deep_2019, ostertag_long-term_2023, cui_survival_2022, liu_cascaded_2023}. Other studies addressed the missing modality problem by designing specific loss functions that take into account only available modalities \citep{gao_deep_2021, xue_ai-based_2024, kawahara_seven-point_2019, nguyen_clinically-inspired_2024, taleb_contig_2022}, sometimes employing a reconstruction loss to reconstruct them \citep{cui_multi-modal_2022}. Interestingly, some studies used learnable embeddings as placeholders for missing modalities \citep{chen_predicting_2024} or directly utilized models, such as transformers, capable of handling input sequences of arbitrary lengths \citep{zhou_uncertainty-aware_2024}, allowing the model to manage cases with missing or incomplete modalities effectively. 

\subsection{Validation}
An observation on the validation of multimodal AI models is that most studies (82\%) are limited to an internal validation scheme. Despite the common use of public datasets, these are often employed as the sole training data rather than as external validation. Another important component in validating multimodal models is the comparison with a unimodal baseline. To provide a broadly representative analysis of the performance gains by multimodal data integration, we equally sampled 3-4 papers per organ system, resulting in a subset of 48 papers that clearly described a multimodal vs unimodal baseline comparison. On average, these multimodal models obtained a 6.2 percentage point increase in AUC, which aligns with the 6.4 percentage points improvement reported previously by \citet{kline_multimodal_2022}. From the 432 papers studied, 72\% of the papers mention an improvement over unimodal models in an ablation experiment. In contrast, only 5\% of the papers did not find any notable improvement by including multiple modalities, and 22\% of papers did not report any comparison with a unimodal baseline. Despite these encouraging findings, it should be noted that these improvements are rarely tested for significance, leading to a potentially optimistic bias.

\section{Clinical applications}\label{sssec:Clinical applications}
Subdividing multimodal AI research papers based on the applied organ system reveals significant variations in the development of multimodal models. The distribution of the papers among the systems is shown in Figure \ref{fig:fig_distribution_medical_disciplines_tasks}a. Interestingly, some studies evaluated their models into more than one system and, therefore, are reported in a separate \textit{Multiple Systems} category (n=27). Papers that did not align with any category were grouped into a \textit{Miscellaneous} category (n=15). Below, we summarize the key contributions in each area.

\subsection{Nervous system (n=122)}
The predominant focus in the nervous system has been on the diagnosis and disease progression of neurodegenerative disorders, such as Alzheimer's disease (n=47), with a few studies focusing on Parkinson's disease (n=9). The second primary focus is cancer diagnosis and survival prediction (n=22). Other studies focused on the diagnosis of autism (n=6), stroke (n=9), and mental disorders, such as schizophrenia (n=6) and depression (n=4). The analyses of these papers showed that most studies in this area utilize MRI as the primary modality, integrating either clinical data (n=43) or 'omics data (n=18), or all three (n=8). Other studies combined pathology with MRI (n=6) or 'omics (n=5), or CT with clinical variables (n=10).

A significant number of studies utilized publicly available datasets (n=115), with the Alzheimer’s Disease Neuroimaging Initiative (ADNI) dataset being the most frequently used (n=44), followed by The Cancer Genome Atlas (TCGA)(n=11), the 'Computational Precision Medicine Radiology-Pathology challenge on brain tumor classification' dataset (CPM-RadPath) (n=6) and Brain Tumor Segmentation Challenge (BraTS) (n=3). Only a limited number of studies conducted external validation of their models (n=15), while the majority employed cross-validation (n=63) and internal validation (n=40). Integrating multiple modalities has enhanced performance in most of these studies (n=83), underscoring the value of incorporating additional data sources that capture information not evident in a single modality alone.

Some studies stood out for their system design and development, with several focusing on glioma grading, such as \citet{li_augmented_2020}, which proposed a patient-wise feature transfer model that learns the relationship between radiological and pathological images. Notably, their model enables inference using only radiology images while linking the prediction outcomes directly to specific pathological phenotypes. Another interesting example is \citet{cui_multi-modal_2022}, which integrates histological images and genomic data and can handle patients with partial modalities by leveraging a reconstruction loss computed on the available modalities. Interestingly, \citet{cui_survival_2022} combined four different modalities, i.e., radiology images, pathology images, genomic, and demographic data, to predict glioma tumor survival and reached a c-index of 0.77 on a 15-fold Monte Carlo cross-validation.

Several studies focused on Alzheimer's prediction, including \citet{liu_cascaded_2023}, that is the only study both handling missing data and externally validating its findings, employing multiple transformer architectures to integrate imaging data and clinical variables and reaching an AUC value of 0.984 on an external validation composed of 382 patients. \citet{pelka_sociodemographic_2020} proposed a unique type of early fusion where sociodemographic data and genetic data are branded on the MRI scan and used to develop a model able to diagnose early and subclinical stages of Alzheimer’s Disease. \citet{lei_alzheimers_2024} developed a novel framework for AD diagnosis that fuses four modalities: genomic, imaging, proteomic, and clinical data, validating the proposed method on the ADNI dataset and outperforming other state-of-the-art multimodal fusion methods. 

The most extensive study of this category is the one of \citet{xue_ai-based_2024} that presented an AI model that integrates diverse data, including clinical data, functional evaluations, and multimodal neuroimaging, to identify 10 distinct dementia etiologies, across 51,269 participants from 9 independent datasets, even in the presence of incomplete data.

\subsection{Respiratory system (n=93)}
Research in the respiratory system predominantly focused on diagnosis (n=10), survival (n=9), treatment response prediction (n=3), and progression (n=2) of cancer, and on diagnosis (n=19), survival (n=3) and disease progression (n=2) of Covid-19. Across these studies, the common strategy involves combining clinical variables with either X-ray or CT imaging of the thorax (n=18 and n=27, respectively). Additionally, a subset of papers (n=11) developed vision-language models by integrating chest X-rays with clinical reports, specifically for X-ray report generation. The majority of the studies regarding respiratory diseases employed public datasets, such as the MIMIC-CXR dataset (n=9), The Cancer Genome Atlas (TCGA) (n=5), and the National Lung Screening Trial (NLST) dataset (n=3). In this category, many studies performed better than single-modality models when performing multimodal integration (n=69). However, the proportion of studies employing external validation is still limited (n=15). 

Among the studies of the respiratory category that stood out, \citet{keicher_multimodal_2023} proposed a multimodal graph‑based approach combining imaging and non‑imaging information to predict COVID-19 patient outcomes. Thanks to the employed attention mechanism, the model learns to identify the neighbors in the graph that are the most relevant for the prediction task, providing insight into the decision process. 
\citet{gao_reducing_2022} outperformed state-of-the-art models on three different external validation sets in predicting the risk of indeterminate pulmonary nodules with a multimodal approach combining CT imaging and clinical variables, where most studies do not perform external validation. 
\citet{gao_deep_2021} proposed a 'multi-path multimodal missing' network, a system integrating multimodal data, including clinical data, biomarkers, and CT images, that can be trained end-to-end with even incomplete data types. The resultant model can make predictions using even a single modality. Cross-validation and external validation showed that combining multiple modalities significantly improves performance in predicting lung cancer risk compared to using a single modality. 
\citet{wang_multi-modality_2024} proposed a novel lung cancer survival analysis framework using multi-task learning, incorporating histopathology images and clinical information, reaching a cross-validation C-index of 0.73 that outperformed traditional methods.
\citet{kumar_deep-learning-enabled_2023} combined X-ray with clinical information to develop a multimodal fusion approach to detect lung disease and developed two multimodal network architectures based on late and intermediate fusion, showing better performances with the latter. 
\citet{lopez_reducing_2020} examined all three multimodal fusion types, early-intermediate-late, with deep learning-based techniques for classifying several chest diseases using radiological images and associated text reports, demonstrating the potential of multimodal fusion methods to yield competitive results using less training data than their unimodal counterparts. 

\subsection{Digestive system (n=43)}
In the domain of the digestive system, most studies focused on the diagnosis of malignancies (n=25), with fewer addressing survival (n=10), treatment response (n=3), and disease progression prediction (n=2). The primary malignancies investigated include colorectal (n=12) and liver (n=10) cancers, with some studies also focusing on the stomach, esophagus, and duodenum (n=9). No modality combination was dominant in this field, though clinical variables (n=31) and histopathology slides (n=17) were commonly included. Integrating multimodal data improved performance in 38/39 papers compared to unimodal approaches. Despite being the category with the highest proportion of externally validated studies (n=11), external validation remains limited. Although quite some studies employed publicly available datasets (n=16), including mainly TCGA (n=11), these datasets were often used for training rather than external validation. 

Several studies were particularly noteworthy. \citet{cui_diagnosing_2024} developed a multimodal AI model using both endoscopic ultrasonography images and clinical information to distinguish carcinoma from noncancerous lesions of the pancreas and tested this model in internal, external, and prospective datasets. They also evaluated the assisting potential of the model in a crossover trial. Their results showed that AI assistance significantly improved the diagnostic accuracy of novice endoscopists and that the additional interpretability information helped reduce skepticism among experienced endoscopists.
\citet{chen_predicting_2024} introduced a deep learning model combining three modalities, radiology, pathology, and clinical data, to predict treatment responses to anti-HER2 therapy or anti-HER2 combined immunotherapy in patients with HER2-positive gastric cancer. This model can manage missing modalities thanks to the incorporation of learnable embeddings which, as substitutes for the modality-specific features, are concatenated to the features of the existing modalities to generate the inter-modal fused feature. 
\citet{zhou_deep_2023} developed and externally validated a multimodal model to predict treatment response to bevacizumab in liver metastasis of colorectal cancer patients using three modalities, PET/CT features, histopathology slides, and clinical data, reaching an AUC of 0.83 in the external validation set. 

\subsection{Reproductive system (n=43)}
The reproductive domain encompasses studies focused on the diagnosis and prognosis prediction of malignancies, mainly in the breast (n=32), followed by the prostate (n=5), ovaries, cervix, and placenta (n=5). Unlike other domains, there was no predominant combination of modalities; clinical variables with MRI (n=9) or histopathology (n=6), omics data with histopathology (n=5) or clinical data (n=3) were explored with similar frequency. As in other categories, most studies demonstrated improved performances when integrating multiple modalities (n=33). However, only a few studies (n=9) performed external validation. To train and internally validate their models, some studies employed public datasets, such as the TCGA-BRCA dataset (n=11).

Among interesting studies, \citet{mondol_biofusionnet_2024} presented a deep learning framework that fuses image-derived features with genetic and clinical data to perform survival risk stratification of ER+ breast cancer patients. They compared their model to six state-of-the-art models, such as MultiSurv, TransSurv, and MCAT, demonstrating an improvement in the AUC value ranging from 0.13 to 0.37.
\citet{holste_end--end_2021} evaluated various methods for fusing MRI and clinical variables in breast cancer classification and achieved an AUC value of 0.989 on an internal validation cohort of 4909 patients, one of the largest validation cohorts we encountered in this domain. \citet{wang_predicting_2023} explored multiple instance learning techniques to combine histopathology with clinical features to predict the prognosis of HER2-positive breast cancer patients and found that simpler modality fusion techniques, such as concatenation, were ineffective in enhancing the model's performance.

\subsection{Sensory system (n=25)}
In the sensory system, most multimodal models have focused on ophthalmology (n=23), with the remaining studies addressing otology (n=2). There is a distinct focus in ophthalmology on glaucoma (n=7), followed by retinopathy (n=4). Three modalities are the most used in these studies: optical coherence tomography, color fundus photography, and clinical data in different combinations. A high proportion of studies achieved better results employing a multimodal model than unimodal ones (n=19), but only a small number of these papers included external validation (n=3).

We want to highlight \citet{nderitu_predicting_2024}, who developed a deep learning system to predict one-, two-, and three-year progression of diabetic retinopathy using risk factors and color fundus images, employing more than 160000 eyes for the development of the model and around 28000 and 7000 eyes to internally and externally validate their findings.
\citet{zhou_uncertainty-aware_2024} employed a multimodal approach that combined four imaging modalities with free-text lesion descriptions for uncertainty-aware classification of retinal artery occlusion. Interestingly, their model can also process incomplete data thanks to the Transformer architecture, which is designed to handle input of arbitrary lengths. 

\subsection{Integumentary system (n=24)}
All studies within the integumentary category focused on diagnosing skin lesions, primarily through the fusion of dermatoscopic images and clinical variables (n=12). Most of the studies (n=14) compared multimodal approaches to unimodal baselines and reported improved performance with adding extra modalities, suggesting that factors such as lesion location and patient demographics contribute valuable information. Notably, one paper conducted prospective validation of its findings \citep{zhu_deep_2024}. Only four remaining studies carried out external validation, while all others relied on internal or cross-validation approaches. In contrast to other systems, several publicly available datasets are readily accessible. The most used one is the dataset coming from the ISIC challenge (n=9), followed by HAM10000 (n=4), Seven-point Criteria Evaluation (SPC) (n=3). 

Among the most interesting examples of this category is \citet{zhou_pre-trained_2024}, who presented SkinGPT-4, an interactive dermatology diagnostic system based on multimodal large language models and trained on pairs of images and textual descriptions. SkinGPT-4 was evaluated on 150 real-life cases in collaboration with board-certified dermatologists. This system allows users to upload their skin photographs for diagnosis, enabling it to assess the images autonomously, identify the characteristics and classifications of skin conditions, conduct in-depth analyses, and offer interactive treatment recommendations. Other studies include \citet{tang_fusionm4net_2022} that proposed two fusion schemes to efficiently integrate dermatoscopic images, clinical photographs, and clinical variables for the diagnosis of eight types of skin lesions, and \citet{zhu_deep_2024} that demonstrated that their multimodal model, which incorporated clinical images and high-frequency ultrasound, performed on par or better than dermatologists in diagnosing seventeen different skin diseases. 

\subsection{Cardiovascular (n=20)}
Research in the cardiovascular domain focused exclusively on diagnosis (n=15), with some studies also concentrating on survival, treatment response, and disease progression prediction (n=4). In all but three studies, the proposed models incorporated clinical variables with a second modality, often in conjunction with radiology imaging (n=10). Most studies obtained better results when comparing unimodal models with multimodal ones(n=15). However, in this case, only three studies externally validated their results. Some publicly available datasets were employed, including MIMIC, JSRT Database, Montgomery County X-ray Set, and data from the Gene Expression Omnibus (GEO) and UK Biobank. 

One interesting example in this category is the study of \citet{v_novel_2024} that introduced an Attention-Based Cross-Modal (ABCM) transfer learning framework to predict cardiovascular disease, merging diverse data types, including clinical records, medical imaging, and genetic information through an attention-driven mechanism. They reached an AUC value of 0.97 on the validation set, significantly surpassing traditional single-source models.

\subsection{Urinary system (n=11)}
In the urogenital system, the majority of studies focused on the diagnosis of kidney (n=8) disorders, with fewer addressing bladder (n=2) and adrenal gland (n=1) conditions. These multimodal studies were employed for diagnosis (n=7) or survival prediction (n=4) predominantly on oncological malignancies (n=9) with fewer on renal artery stenosis and aiming at performing report generation. Most studies employed CT images combined with a second or third modality (n=9), such as clinical data, histopathology images, and omics. Eight studies demonstrated improved performance with multimodal models compared to unimodal ones. Similarly to other systems, only a small number (2/11) externally validated their results, while most used internal or cross-validation approaches. A few public datasets are available for this category: The Cancer Genome Atlas (TCGA) and The Cancer Imaging Archive (TCIA).

One study focused on renal diseases developed a cross-modal system to create a prognostic model for clear renal cell carcinoma \citep{ning_integrative_2020}. This model integrates deep features extracted from computed tomography and histopathological images with eigengenes derived from genomic data, demonstrating a significant ability to stratify patients into high- and low-risk categories for disease progression.

\subsection{Musculoskeletal system (n=9)}
In the musculoskeletal system, studies focused on diagnosing bone (n=6) diseases, with fewer addressing teeth and muscle conditions, focusing on diagnosing pathological gait, osteoarthritis, deep caries and pulpitis, bone fractures and aging, and soft tissue sarcoma. These studies mainly employed radiology images with clinical data (n=6) or unstructured text data (n=2). Most studies (n=6) demonstrated improved performance with multimodal models compared to unimodal ones. However, no studies externally validated their results, mainly employing internal validation approaches. This category used two public datasets from the Osteoarthritis Initiative (OAI) and the Pediatric Bone Age Challenge. 

\citet{li_hgt_2023} employed the most extensive validation set of the category, including more than 24000 samples, to diagnose Prosthetic joint infection from CT scan and patients' clinical data using a Unidirectional Selective Attention mechanism and a graph convolutional network and reaching an AUC value of 0.96. \citet{schilcher_fusion_2024} included a multicentre dataset of 72 Swedish radiology departments to develop a multimodal model based on X-ray imaging and clinical information for detecting atypical femur fractures.

\subsection{Multiple systems (n=27)}
This category includes papers that evaluated the multimodal system on multiple organs belonging to different systems. All these studies benefited from available public datasets, confirming that publicly available data is essential for developing and assessing multimodal AI systems. Most of the studies in this category evaluated the performance of their models across diseases affecting up to five organs (n=20). Four studies were distinguished by the extensive range of organs on which they assessed their models. 
\citet{azher_assessment_2023} developed an interpretable multimodal modeling framework that combines DNA methylation, gene expression, and histopathology for the prognostication of eight types of cancers.
\citet{chen_pan-cancer_2022} developed a multimodal deep learning model to jointly examine pathology whole-slide images and molecular profile data from 14 cancer types to predict outcomes and discover prognostic features correlating with poor and favorable outcomes. Notably, a vital explainability component was incorporated with heatmaps for histopathology and SHAP values for genomic markers.
\citet{cheerla_deep_2019} constructed a multimodal neural network-based model to predict the survival of patients for 20 different cancer types using clinical data, mRNA expression data, microRNA expression data, and histopathology whole slide images (WSIs). 
Lastly, \citet{vale-silva_long-term_2021} presented a multimodal deep learning method for long-term pan‑cancer survival prediction that was applied to data from 33 different cancer types, the highest number of organs of the category. These four studies demonstrated the reliability of their models on a wide range of cancer types to do that interestingly, all four benefited from the publicly available data of TCGA.

\subsection{Miscellaneous (n=15)}
This category includes papers that do not fit into the categories above, such as diagnosis, recurrence, and survival prediction of patients with thyroid carcinoma, prediction of Type II diabetes, severe acute pancreatitis, or immunotherapy response for diffuse large B-cell lymphoma, abnormality detection from chest images, fetal birth weight prediction, and other specific tasks that did not fit the categories above. However, some studies provide novel or unique contributions to multimodal model development. Some interesting examples include \citet{kim_prediction_2023} that proposed a model employing an autoencoder with multiple encoders to extract comprehensive features from hormonal and pathological data and demonstrated that the proposed model significantly improves performance when compared to unimodal models when predicting the recurrence probability of thyroid cancer patients. \citet{lee_prediction_2024} explored unsupervised learning by combining histopathology and clinical data, followed by knowledge distillation to derive a unimodal histopathology model for predicting immunochemotherapy response of patients with Diffuse large B-cell lymphoma. 
\citet{khader_multimodal_2023} developed a transformer-based neural network architecture that integrates multimodal patient data, including both imaging and non-imaging, to diagnose up to 25 pathologic conditions and showed that the multimodal model significantly improves diagnostic accuracy when using both chest radiographs and clinical parameters compared to imaging alone and clinical data alone. Finally, \citet{cai_pre-trained_2023} presented a pre-trained multilevel fusion network that combines Vision-conditioned reasoning and Bilinear attentions to enhance feature extraction from medical images and questions. By incorporating Contrastive Language-Image Pre-training (CLIP) and stacked attention layers, their model reduces language bias and improves accuracy. Experiments on three benchmark datasets showed that the proposed model surpasses state-of-the-art models.

\section{Towards clinical implementation} \label{ssec:road_to_the_clinic}
In addition to reviewing the landscape of multimodal AI models in research, we investigated to what extent these potential performance boosts can be achieved by clinicians today. To this end, we searched the \href{https://www.fda.gov/medical-devices/software-medical-device-samd/artificial-intelligence-and-machine-learning-aiml-enabled-medical-devices}{FDA database of AI/ML-enabled medical devices} and the \href{https://healthairegister.com/}{Health AI register} for multimodal AI models that obtained either FDA- or CE-clearance. Since these databases contain 950 and 213 entries, we performed an automated preselection of models that may incorporate multiple modalities. Specifically, for each entry in the FDA database, we automatically retrieved the publicly available 510(k) premarket notification summary and searched for any mention of "multi-modal" or "multimodal". This resulted in 38 hits, all manually inspected for the same inclusion criteria used in the literature search. Although one product was found to fit the multimodality criterion by incorporating EEG, neurocognitive test scores, and clinical symptoms to assess the need for CT imaging after potential structural brain injury \citep{hanley_emergency_2017}, no deep neural networks were involved, which would place this out of scope for the current review. 

A similar strategy was employed for entries in the Health AI register. When available, a published validation study was retrieved and scanned for occurrences of the word "multimodal" or "multi-modal." Although a validation study could be retrieved for 132/213 products, none of these studies returned a hit for our multimodal keywords. Combining our findings from the Food and Drug Administration (FDA) database and AI health register, the promising multimodal AI models showcased in this review have yet to make their way to the clinic. 

However, despite the lack of FDA— or CE-certified multimodal AI models, we identified two papers during our literature search that demonstrated potential for transitioning multimodal AI models from research to clinical practice.
First, \citet{esteva_prostate_2022} describes a multimodal model that integrates digital pathology slides and clinical variables for risk stratification in prostate cancer. Specifically, the model employs a ResNet50 pretrained in a self-supervised manner to extract a feature vector from the pathology slides and subsequently concatenates this image feature vector with a clinical variable vector for the final prediction. Compared to the unimodal imaging model, the multimodal obtained a notably higher AUC of 0.837 vs. 0.779 for predicting distant metastasis after 5 years. In addition, this model was externally validated in a phase 3 trial (NRG/RTOG 9902) and demonstrated significant and independent association with prognostic factors compared to current methods such as the National Comprehensive Cancer Network (NCCN) high-risk features \citep{ross_external_2024}. Given this favorable head-to-head performance comparison with current clinical predictive methods, it has since been incorporated into the NCCN guidelines. Although this model is not FDA-certified, clinicians in the United States of America can order it as part of the regular prognostic workup of their patients. 

Second, \citet{lee_development_2022} focused on developing and validating a multimodal prognosis and treatment selection model in COVID-19 patients. This multimodal model employed a previously validated and CE-approved chest X-ray abnormality detection model \citep{nam_development_2021} to infer imaging features from chest X-rays and fused this with clinical and laboratory data to predict prognosis and the required interventions for patients with COVID-19. Retrospective validation on 2282 COVID-19 patients from 13 medical centers demonstrated that the multimodal model significantly outperformed the unimodal imaging model with an AUC of 0.854 vs. 0.770, respectively. A feature importance analysis revealed that, among others, clinical variables such as age and dyspnea played an essential role in the multimodal model's prediction. Although the multimodal model is unavailable for clinicians today, this study proved that current CE-certified unimodal AI models can be incorporated into a multimodal model pipeline to improve performance. 

As briefly mentioned in the introduction, medical data is siloed, with each discipline having its own data storage systems. Examples are the Picture Archiving and Communication System (PACS) for radiology, Image Management System (IMS) for pathology, and Electronic Health Records (EHR) for clinical information. Of course, this is an issue for data curation and collection but also a huge issue for implementation, as AI would have to run on data sourced from different systems that do not interoperate. 

Additionally, previous studies have demonstrated that by utilizing only a little background information about participants, an adversary could re-identify those in large datasets \citep{narayanan_robust_2008}. As the amount of multimodal data collected per patient increases, phenotyping accuracy improves, and significant privacy concerns rise, as the richer data can lead to the re-identification of individuals within large datasets. Managing these privacy issues is crucial to protect patient confidentiality while benefiting from the enhanced insights provided by comprehensive multimodal data.

Lastly, the development of explainable AI (XAI) is of importance to improve trustworthiness by addressing the limitations of AI's "black box" nature \citep{pahud_de_mortanges_orchestrating_2024}. Explainable AI ensures that end-users can understand and trust the AI's decision-making process, which is crucial for the widespread adoption and usability of these technologies in clinical practice \citep{kline_multimodal_2022}. Prioritizing these strategies will help bridge the gap between research and practical, commercially available multimodal AI solutions in healthcare.

\section{Discussion}
The exponential growth in multimodal AI models highlights the recent research efforts of multimodal data integration in healthcare. Although multimodal AI development poses unique challenges, the increasing research output in the field will inevitably lead to overcoming some of these challenges. Importantly, in line with a previous review \citep{kline_multimodal_2022}, our review revealed that integrating multiple data modalities leads to notable performance boosts. In the remainder of this section, we will highlight key takeaways of our analysis and provide recommendations for future research directions. 

The general state of the multimodal medical AI development field was summarized in section \ref{overview_modalities}. This overview revealed significant disparities in multimodal AI development across various medical disciplines, tasks, and data domains. Notably, multimodal AI development commonly finds applications in the nervous and respiratory systems. Conversely, applications in the musculoskeletal system or urinary system were scarce. Similar disparities are evident in the realm of data modality combinations. The integration of radiology imaging data and text data greatly outnumbered other combinations and was investigated in almost half of all papers in this review.
On the other hand, integrating radiology and pathology data into multimodal frameworks seems to present more significant challenges. Furthermore, substantial variations were observed in the model's application for a given medical task. Most models were aimed at automated diagnosis, whereas prediction of disease progression or survival was less common. 

Irrespective of the specific medical discipline or application, these discrepancies consistently point to a common factor: high-quality public datasets' availability (or lack thereof). The increased complexity associated with curating multimodal datasets, often requiring data collection from diverse sources (e.g., electronic health records or physical pathology slides) and departments (e.g., radiology and pathology), currently presents a significant bottleneck in model development. However, collecting, curating, and harmonizing detailed, diverse, and comprehensive annotated datasets representing various demographic factors are essential \citep{acosta_multimodal_2022}. The richness and quality of such data are paramount for training effective AI models, as they ensure robustness, generalizability, and accuracy in real-world applications. This process, however, faces challenges such as high costs associated with elaborate patient characterization and longitudinal follow-up, which escalate with increasing participant numbers, making automated data collection methods necessary to remain financially sustainable \citep{acosta_multimodal_2022}.

However, this challenge also presents an opportunity for researchers and clinicians to make impactful contributions to the field, as establishing new multimodal datasets can substantially accelerate AI development in a specific domain. For instance, the publicly available ADNI (Alzheimer's Disease Neuroimaging Initiative) dataset has facilitated the development of more models for Alzheimer's diagnosis (n=45) than all studies combined in the musculoskeletal and urinary system. Apart from enabling domain-specific models, these public datasets can stimulate research tackling general challenges associated with multimodal AI development. When datasets encompass as many as four distinct modalities (i.e. CT imaging, pathology slides, genomic data and text data), as seen in the TCGA-GBM dataset (The Cancer Genome Atlas Glioblastoma Multiforme Collection), this allows extensive investigations on the effects of missing data, model uncertainty, and model explainability with various data modalities \citep{cui_survival_2022, braman_deep_2021, hao_page-net_2020}. Hence, we believe that a community effort towards multimodal dataset development can have a strong propelling effect on the progress of the field. 

The technical design choices associated with processing and fusing data from diverse modalities were covered in sections \ref{sssec:architecture choices} and \ref{sssec:Missing data handling}. Our analysis revealed that most studies prioritize the development of effective modality fusion methods over designing novel encoder architectures for optimal processing of different modalities. This was primarily evidenced by the comparatively large proportion of studies employing pretrained encoders for feature extraction from other modalities, after which the main contribution consisted of a novel, generally intermediate-level, fusion method. We posit that this observed focus on fusion methods rather than data encoders stems from the nature of multimodal datasets, which are typically smaller in scale than their unimodal counterparts. By leveraging encoders pre-trained on large unimodal datasets and occasionally fine-tuning them on the available multimodal data, researchers can achieve robust performance despite the relative scarcity of multimodal data \citep{wang_enabling_2023, zhang_knowledge-enhanced_2023}.

Moreover, the great diversity of modalities in the medical field with their respective intricacies seems to warrant a thorough investigation of optimal data fusion methodologies. A robust fusion method is critical for reliable model predictions, especially in light of the potential absence of specific modalities during inference. The emergence of (medical) foundational models \citep{moor_foundation_2023}, which are generally multimodal, is expected to accelerate research in optimal data fusion further. Since these foundational models can further improve the availability of pre-trained solid encoders, this may shift the research focus to designing optimal feature fusion methods. Where we initially saw the main focus on late fusion techniques, where existing unimodal models could straightforwardly be combined with a structured non-imaging modality, newer work moves towards earlier fusion, with most early-fusion papers, which would allow more cross-modality learning, appearing in the past two years. Early fusion has the theoretical advantage that information from the other modalities can improve each modality's encoder, thus creating a $1+1 > 2$ scenario. However, this is underexplored in the current body of work, and it indicates a promising direction for future research.

Similarly, the development of stronger data encoders may also help tackle the challenge of handling missing data during training and inference. Although the majority of studies in this review discarded entries with missing modalities in their dataset, some innovative approaches leveraged strong encoders to generate learnable embeddings for missing modalities \citep{chen_predicting_2024} or employed encoders with flexible input lengths to handle varying amounts of data \citep{zhou_uncertainty-aware_2024}. 

The clinical applicability of current multimodal AI models, as discussed in subsection \ref{ssec:road_to_the_clinic}, reveals that the readiness of these models for clinical implementation lags behind that of unimodal models. We believe this is influenced by the same challenges present in commercializing unimodal AI models, but amplified by the multimodal nature of these models. Although unimodal models require strong, preferably external, validation before deployment, this is even more prudent for multimodal AI models. However, obtaining cross-departmental multimodal data for external validation can be challenging. In addition, since multimodal models consume larger amounts of data during deployment, there may be a greater risk for model drift when one of the data modalities changes. Although extensive validation can investigate the effect of slight modality changes, such as using slightly different staining in a multimodal model incorporating histopathology imaging data, this may lead to a combinatorial explosion when fusing more data modalities. A more optimistic foreseeable future could be that multiple modalities can serve a stabilizing purpose, where a noisy additional modality is unlikely to influence the final prediction adversely. However, we note that fusion of more than two modalities was encountered in only 59/432 (14\%) papers, indicating that more evidence is needed to corroborate or disprove any of the aforementioned scenarios. Another amplified challenge in commercializing multimodal AI models is the call for explainable AI. Although this is already challenging for unimodal models, explainability in multimodal AI models likely requires explainability for each component (i.e., Shapley values for tabular data), but perhaps more importantly, some degree of explainability of how information from different sources is combined. Such an explainable multimodal AI model may also alleviate the aforementioned risk of performance degradation under data drift for one of the data modalities. Given these challenges, we believe that developing public multimodal datasets, preferably incorporating more than two modalities, will be an essential stepping stone for advancing the field.

\section{Conclusion}
In conclusion, this review provides one of the most comprehensive overviews of multimodal AI development, spanning various medical disciplines, tasks, and data domains. Although substantial evidence exists that multimodal AI models will incur significant performance boosts by taking a broader view of the patient, their development poses novel challenges. We hope this review elucidated some of these challenges, but more importantly, potential solutions to guide the field in the coming years. 

\section*{CRediT authorship contribution statement}

\textbf{Daan Schouten:} Conceptualization, Methodology, Formal Analysis, Data Curation, Visualization, Writing - original draft. \textbf{Giulia Nicoletti:} Conceptualization, Methodology, Formal Analysis, Data Curation, Visualization, Writing - original draft. \textbf{Bas Dille:} Conceptualization, Methodology, Formal Analysis, Data Curation, Visualization, Writing - original draft. \textbf{Catherine Chia:} Conceptualization, Methodology, Formal Analysis, Data Curation, Visualization, Writing - original draft. \textbf{Pierpaolo Vendittelli:} Methodology, Formal Analysis, Data Curation, Visualization, Writing - original draft. \textbf{Megan Schuurmans:} Methodology, Formal Analysis, Data Curation, Visualization, Writing - original draft. \textbf{Geert Litjens:} Conceptualization, Writing - Review \& Editing, Supervision. \textbf{Nadieh Khalili:} Conceptualization, Writing - Review \& Editing, Supervision.

\section*{Declaration of competing interest}

The authors declare that they have no known competing financial interests or personal relationships that could have appeared to influence the work reported in this paper.

\section*{Data availability}

No data was used for the research described in the article.

\section*{Acknowledgments}

This research has been funded by the Dutch Research Council, the European Union Horizon Europe Program, and the Hanarth Fund. We acknowledge Mr. H. J. M. Roels's financial support through a donation to Oncode Institute.

\section*{Supplementary data} 
Supplementary material related to this article can be found online.

\def\url#1{}
\def\note#1{}
\bibliographystyle{bibliography/model2-names.bst}
\bibliography{bibliography/Included_paper_list}

\end{document}